\documentclass{article}

\PassOptionsToPackage{numbers, compress}{natbib}
\usepackage[preprint]{neurips}

\usepackage[utf8]{inputenc} % allow utf-8 input
\usepackage[T1]{fontenc}    % use 8-bit T1 fonts
\usepackage{hyperref}       % hyperlinks
\usepackage{url}            % simple URL typesetting
\usepackage{booktabs}       % professional-quality tables
\usepackage{amsfonts}       % blackboard math symbols
\usepackage{nicefrac}       % compact symbols for 1/2, etc.
\usepackage{microtype}      % microtypography
\usepackage{xcolor}         % colors

\usepackage{amssymb} % use thicksim
\usepackage{graphicx} % control formula size
\usepackage{amsmath} % use /operatorname
\usepackage{multirow}       % multirows
\usepackage{tabularx}       % tablular with linewidth
\usepackage{colortbl}
\usepackage{color,soul}

\setul{0.5ex}{0.3ex}
\usepackage{utfsym} % use ✅

\title{VideoCap-R1: Enhancing MLLMs for Video Captioning via Structured Thinking}

\author{Desen Meng$^{1*}$, Rui Huang$^{1*}$, Zhilin Dai$^{1}$, Xinhao Li$^{1,2}$, Yifan Xu$^{1}$, Jun Zhang$^{1}$ \\ \textbf{Zhenpeng Huang}$^{1}$, \textbf{Meng Zhang}$^{3}$, \textbf{Lingshu Zhang}$^{3}$  \textbf{Yi Liu}$^{\dagger3}$, \textbf{Limin Wang}$^{\dagger1,2}$  \\
    \small$^1$Nanjing University~~~ 
    \small$^2$Shanghai AI Laboratory~~~
    \small$^3$Honor Device Co., Ltd  \\
}

\begin{document}

\maketitle
{
\renewcommand{\thefootnote}%
{\fnsymbol{footnote}}
\footnotetext[0]{\hspace*{-2.1em} $\dagger$ Corresponding authors: lmwang@nju.edu.cn; liuyi26@honor.com \\ 
* Equal contribution. } 
}

% \begin{abstract}
%   The abstract paragraph should be indented \nicefrac{1}{2}~inch (3~picas) on
%   both the left- and right-hand margins. Use 10~point type, with a vertical
%   spacing (leading) of 11~points.  The word \textbf{Abstract} must be centered,
%   bold, and in point size 12. Two line spaces precede the abstract. The abstract
%   must be limited to one paragraph.
% \end{abstract}

\begin{abstract}
  While recent advances in reinforcement learning have significantly enhanced reasoning capabilities in large language models (LLMs), these techniques remain underexplored in multi-modal LLMs for video captioning. This paper presents the first systematic investigation of GRPO-based RL post-training for video MLLMs, with the goal of enhancing video MLLMs' capability of describing actions in videos. Specifically, we develop the \textbf{VideoCap-R1}, which is prompted to first perform structured thinking that analyzes video subjects with their attributes and actions before generating complete captions, supported by two specialized reward mechanisms: a LLM-free think scorer evaluating the structured thinking quality and a LLM-assisted caption scorer assessing the output quality. The RL training framework effectively establishes the connection between structured reasoning and comprehensive description generation, enabling the model to produce captions with more accurate actions. Our experiments demonstrate that VideoCap-R1 achieves substantial improvements over the Qwen2VL-7B baseline using limited samples (1.5k) across multiple video caption benchmarks (DREAM1K: \textbf{+4.4} event F1, VDC: \textbf{+4.2} Acc, CAREBENCH: \textbf{+3.1} action F1, \textbf{+6.9} object F1) while consistently outperforming the SFT-trained counterparts, confirming GRPO's superiority in enhancing MLLMs' captioning capabilities. 
\end{abstract}
\section{Introduction}
\label{sec:Introduction}
Test-time scaling has been proven to effectively enhance the reasoning capabilities of large language models (LLMs), as demonstrated by OpenAI's o1~\cite{jaech2024openaio1}, Deepseek-R1~\cite{guo2025deepseekr1}, and Kimi-1.5~\cite{team2025kimi1.5}, which exhibit strong performance in complex logical tasks such as mathematics\cite{lightman2023let} and coding\cite{jain2024livecodebench}. Notably, Deepseek-R1 showcases the potential of LLMs to develop reasoning abilities without any supervised data, relying solely on pure reinforcement learning with rule-based verifiable rewards. 

Many researchers\cite{meng2025mmEureka,liu2025Visual-rft,shen2025Vlm-r1,feng2025Video-r1,li2025VideoChat-R1,zhou2025r1} have devoted significant efforts to extending Deepseek-R1's paradigm to the multimodal large language models (MLLMs), aiming to improve visual reasoning capabilities. For instance, MM-EUREKA\cite{meng2025mmEureka} focuses on multimodal mathematical tasks with visual inputs, revealing a "visual aha moment" where the model reaffirms its answer by re-perceiving the image. Furthermore, Visual-RFT\cite{liu2025Visual-rft} and VLM-R1\cite{shen2025Vlm-r1} enhance MLLMs' performance in fundamental visual perception tasks, including detection and grounding. For video understanding, Video-R1 \cite{feng2025Video-r1} proposes T-GRPO, explicitly encouraging temporal reasoning, while VideoChat-R1\cite{li2025VideoChat-R1} conducts deeper analyses and comprehensive ablation experiments on video reasoning mechanisms. These works collectively validate the superiority of the GRPO\cite{shao2024deepseekmath} algorithm over supervised fine-tuning (SFT) in specific visual tasks, such as visual question answering\cite{zhao2025mmvu,yang2024thinking}, spatial grounding\cite{yu2016modeling}, and temporal grounding\cite{gao2017tall}. However, they primarily focus on verifiable problems (e.g., math\cite{lu2023mathvista,zhang2024mathverse,he2024olympiadbench}, multiple-choice questions\cite{zhao2025mmvu,yang2024thinking,hu2025video}), leaving open-ended problems like video captioning\cite{chai2025auroracap,xu2024carebench,wang2024tarsier,caba2015activitynet} underexplored. 

\begin{figure}[t]
  \centering
  \includegraphics[width=0.87\linewidth]{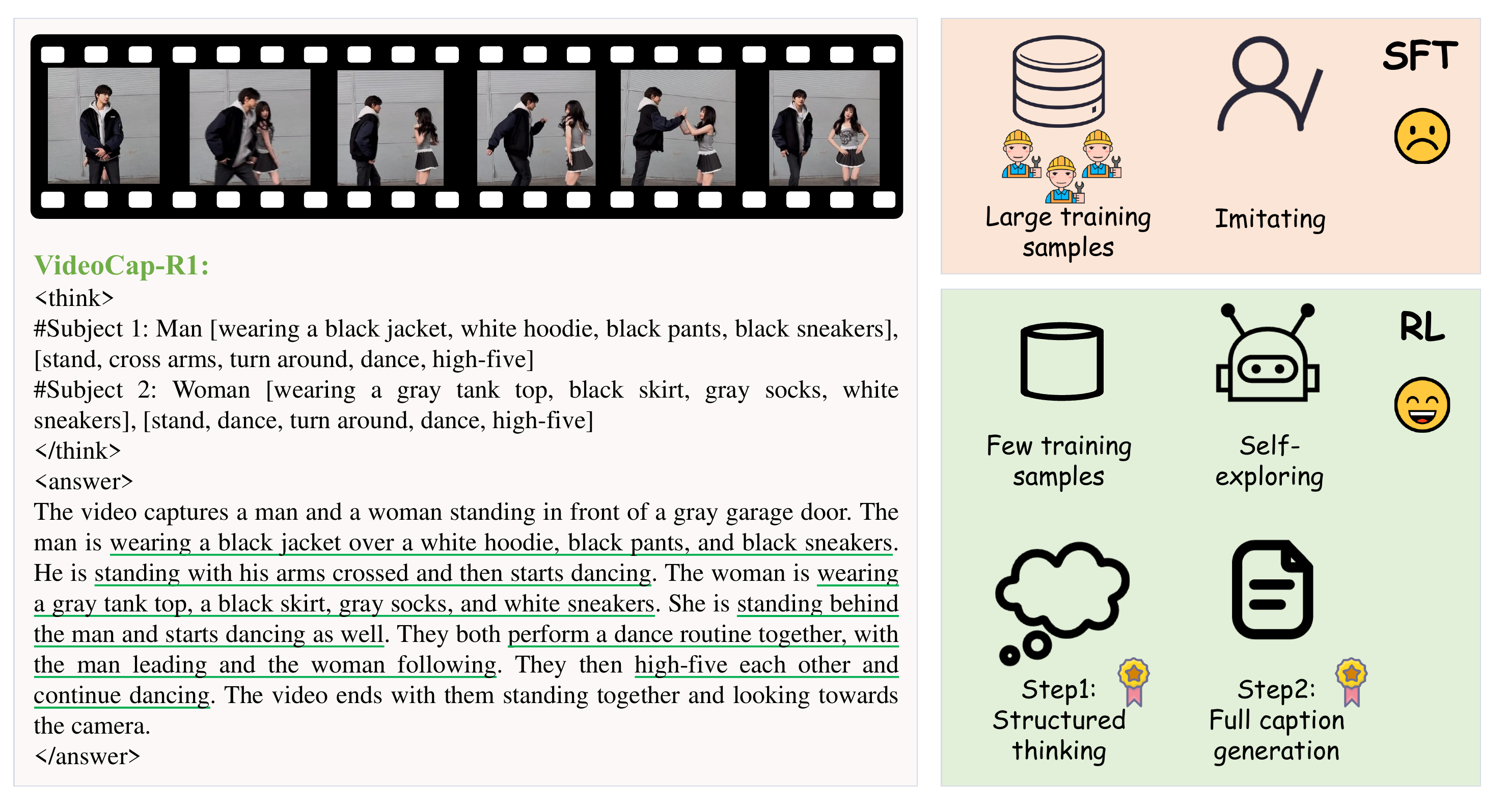}
  \caption{\textbf{Motivation of VideoCap-R1.} SFT requires costly high-quality data and the trained model merely imitates training distributions. VideoCap-R1 instead decomposes captioning into structured thinking and answering phases, optimized via GRPO with dual rewards for thinking and caption. By effectively establishing the connection between structured reasoning and comprehensive description generation, VideoCap-R1 can generate captions with more accurate actions.}
  \label{fig:main_pic}
\end{figure}

In video captioning, most existing approaches rely on manually annotated or commercial model-generated (e.g., GPT-4o\cite{hurst2024gpt}, Gemini-2.0\cite{pichai2024introducinggemini2.0}) high-quality video description datasets for instruction tuning, which is both time-consuming and costly. Inspired by the Chain-of-Thought (CoT) paradigm\cite{xu2024llavacot}, we decompose the captioning task into two sequential steps: first prompting the model to perform structured reasoning that analyzes video subjects, their attributes and actions, then requiring it to synthesize these elements into a complete caption.

We initially constructed an instruction-tuning dataset incorporating this structured reasoning process and performed supervised fine-tuning (SFT) on the baseline model. However, we observed that the model only learned superficial reasoning patterns - it acquired the output format without establishing meaningful connections between the structured reasoning process and final descriptions. We attribute this limitation to the train-inference discrepancy in SFT: during training, the model generates tokens conditioned on ground-truth prefixes, whereas during inference it must rely on its own predictions, resulting in poorer performance when generating captions with structured thinking process compared to direct generation.

To address these limitations, we leverage the recent success of GRPO-based RL post-training strategies, which can provide online rewards for correct reasoning paths. This enables the model to genuinely learn the two-step process of first solving simpler subproblems before generating complete captions. The primary challenge in applying GRPO to video captioning lies in the difficulty of directly comparing generated captions with ground truth for reward assignment. To overcome this, we designed two specialized reward mechanisms: a LLM-free think scorer that evaluates reasoning quality, and a LLM-assisted caption scorer that assesses output quality. Based on this framework, we developed VideoCap-R1, which first identifies key visual elements before generating detailed descriptions, significantly enhancing the baseline model's captioning capability even with limited training samples(1.5k).

Our main contributions are summarized as follows:
\begin{itemize}
    \item We propose a novel structured reasoning process specifically designed for caption generation, where the model first identifies key visual subjects along with their attributes and actions before generating comprehensive descriptions. Under GRPO training, this structured approach demonstrates substantial gains in caption quality.

    \item We present the first successful application of GRPO to open-ended video captioning tasks. Our work introduces two meticulously designed reward mechanisms that jointly assess both the reasoning process and the final caption quality. Based on this framework, we develop VideoCap-R1, which demonstrates consistent performance improvements over baseline models(Qwen2-VL-7B\cite{wang2024qwen2vl}) even with limited training samples(1.5k) across three challenging benchmarks: DREAM-1K(\textbf{+4.4} event F1), VDC(\textbf{+4.2} accuracy), and CAREBENCH(\textbf{+3.1} action F1, \textbf{+6.9} object F1).

    \item Our analysis reveals that SFT only enables models to learn superficial reasoning patterns. This is evidenced by models fine-tuned with structured thinking augmented data underperforming those trained on standard captioning data. In contrast, our GRPO-based RL approach enables the model to develop genuinely beneficial reasoning patterns, outperforming SFT-based counterparts when using identical training datasets, regardless of structured thinking augmentation.
\end{itemize}
\section{Related work}
\label{sec:Related_work}
\textbf{Video caption models.} Video captioning is one of the most fundamental tasks in video understanding. Since video captioning datasets are commonly used in the pre-training phase of multimodal large language models (MLLMs) to align linguistic and visual space, general video MLLMs\cite{zhang2024llavavideo,wang2024qwen2vl,InternVL2.5} typically possess basic video captioning capabilities. The prevailing approach to enhancing MLLMs' video captioning performance involves constructing high-quality video description datasets for instruction tuning. For instance, ShareGPT4Video\cite{chen2024sharegpt4video} designs a differential video captioning strategy, leveraging GPT-4V\cite{achiam2023gpt4} to annotate videos and develop ShareCaptioner-Video. Similarly, Shot2Story~\cite{han2023shot2story20k} and Vript\cite{yang2024vript} employ GPT-4V for video captioning. LLaVA-video \cite{zhang2024llavavideo} introduces a recurrent detailed caption generation pipeline powered by GPT-4o, enabling fine-grained descriptions for videos of arbitrary length. Tarsier2\cite{yuan2025tarsier2} further advances this direction by curating 40 million large-scale video-text pairs for pretraining and 150K human-annotated video descriptions with temporal grounding for instruction tuning. While these specialized video description models (VDCs)\cite{chen2024sharegpt4video,zhang2024llavavideo,yuan2025tarsier2,han2023shot2story20k} excel at generating detailed captions, they predominantly rely on large-scale, manually annotated instruction-tuning datasets, which are costly and time-consuming to produce. In contrast, our work explores training efficiency by leveraging reinforcement learning (RL) to guide the model in reasoning before generating captions. Under the same data budget, our approach outperforms supervised fine-tuning (SFT), demonstrating superior data efficiency.

\textbf{Reinforcement learning for MLLMs.}
Reinforcement learning (RL) is typically applied during the post-training phase of LLMs and has been proven to be critical for mitigating hallucination or enhancing reasoning capabilities. The OpenAI's o1 model\cite{jaech2024openaio1} first demonstrated the significant potential of test-time scaling in improving model reasoning. Subsequently, DeepSeek-R1\cite{guo2025deepseekr1} showed that reinforcement learning with rule-based verifiable rewards could effectively enhance LLMs' performance in mathematical and coding tasks. This approach inspired numerous efforts to extend the R1's paradigm to multimodal domains to improve MLLMs' reasoning abilities. In the video domain, prior work\cite{wang2025timezero,zhao2025R1-omni,li2025VideoChat-R1} has explored the effectiveness of GRPO in tasks such as temporal grounding, sentiment analysis, object tracking, and general visual question answering. However, open-ended tasks like video captioning remain understudied. For instance, VideoChat-R1 \cite{li2025VideoChat-R1}attempted to improve video description quality using event recall as a reward function, but the generated captions remained far from satisfactory. Our work addresses this gap by systematically designing and evaluating reward functions tailored for captioning, successfully adapting GRPO to this task and significantly improving description quality.

\section{Methodology}
\label{sec:Methodology}
\subsection{Preliminary}

\subsubsection{Group Relative Policy Optimization}
Group Relative Policy Optimization (GRPO)\cite{shao2024deepseekmath} is an enhanced variant of Proximal Policy Optimization (PPO)\cite{schulman2017proximalPPO}. GRPO obviates the need for additional value function and uses the average reward of multiple sampled outputs for the same question to estimate the advantage. To be specific, for each question-answer pair $(q,a)$, the old policy $\pi_{\theta_{\mathrm{old}}}$ samples a group of outputs $\{o_1,o_2,\ldots, o_G\}$ and a predefined reward function is used to evaluate these outputs to get their corresponding rewards $\{r_1,r_2,\ldots, r_G\}$. Then the advantage of the $i$-th response relative to other sampled responses is calculated by normalizing the group-level rewards $\{r_1,r_2,\ldots, r_G\}$: $\hat{A}_{i}=\frac{r_i-\operatorname{mean}(\{r_1,r_2,\ldots, r_G\})}{\operatorname{std}(\{r_1,r_2,\ldots, r_G\})}$.
GRPO encourages the model to prioritize the responses with higher advantages within the group by updating the policy $\pi_{\theta}$ using the following clipped surrogate objective:
\begin{equation}
\resizebox{.9\hsize}{!}{$
\begin{aligned}
\mathcal{J}_{\mathrm{GRPO}}(\theta) & =\mathbb{E}_{(q,a)\thicksim\mathcal{D},\{o_i\}_{i=1}^G\thicksim\pi_{\theta_{\mathrm{old}}}{(\cdot|q)}} \\
 & \left[\frac{1}{G}\sum_{i=1}^G\frac{1}{|o_i|}\sum_{t=1}^{|o_i|}\left(\min\left(\frac{\pi_\theta(o_{i,t}\mid q,o_{i,<t})}{\pi_{\theta_{\mathrm{old}}}(o_{i,t}\mid q,o_{i,<t})}\hat{A}_{i},\mathrm{~clip}{\left(\frac{\pi_\theta(o_{i,t}\mid q,o_{i,<t})}{\pi_{\theta_{\mathrm{old}}}(o_{i,t}\mid q,o_{i,<t})},1-\varepsilon,1+\varepsilon\right)}\hat{A}_{i}\right)-\beta D_{\mathrm{KL}}(\pi_{\theta}||\pi_{\mathrm{ref}})\right)\right],
\end{aligned}
$}
\end{equation}
where $\varepsilon$ and $\beta$ are hyper-parameters. $\varepsilon$ is the clipping range of importance sampling ratio and KL divergence is adopted to regularize the policy model, preventing excessive deviation from the reference model. 

\subsection{Caption Reward Modeling}
The reward function plays a pivotal role in determining the optimization direction of reinforcement learning. DeepSeek-R1 adopts a rule-based reward system comprising two primary components: format reward and accuracy reward. Building upon DeepSeek-R1's framework, we introduce novel caption-specific reward functions to guide policy optimization, whose components and implementation details are elaborated as follows.
\subsubsection{Two-step Caption Generation Strategy}
While video captioning is fundamentally a perceptual task requiring comprehensive description of visual elements, it presents unique challenges compared to visual question answering. Unlike VQA tasks\cite{li2024mvbench,fu2024videovideomme} that focus only on question-relevant content, video captioning demands complete coverage of all significant elements within potentially complex video sequences.

Inspired by the Chain-of-Thought (CoT) paradigm\cite{xu2024llavacot, zhang2024improve, thawakar2025llamao1, yao2024mulberry} that decomposes complex tasks into manageable sub-problems, we propose a two-step caption generation strategy. Our approach first requires the model to perform structured reasoning that analyzes and identifies key video subjects along with their attributes and actions in the thinking process and then synthesize these elements into coherent captions in the final outputs, as illustrated in Figure \ref{fig:main_pic}. The training prompt is detailed in the Appendix \ref{training_prompt}. We employ format reward as Deepseek-R1 to ensure the model adheres to this format. The two-step caption generation strategy mirrors compositional writing, where one first outlines key points before developing complete paragraphs. Our experimental results confirm that this explicit reasoning framework significantly enhances the model's capability to describe actions and events in videos.

\subsubsection{LLM-Free Think Scorer}
To effectively establish the connection between the key elements identified during the thinking process and the final caption outputs, we implement a LLM-free think scorer for the intermediate thinking stage. We extract subject names, attribute lists, and action lists through regular expression matching and compute corresponding precision and recall metrics against ground truth annotations. 

Formally, let the model predict $N$ entities, each containing a name $\text{name}_I^p$, an attribute list $\text{attr}_I^p$, and an action list $\text{act}_I^p$ ($1\le I\le N$), while the ground truth contains $M$ entities with corresponding $\text{name}_J^g$, $\text{attr}_J^g$, and $\text{act}_J^g$ ($1\le J\le M$). For each predicted action list $\text{act}_I^p=\{p_i\}_{i=1}^{n}$ and its corresponding ground truth action list $\text{act}_J^g=\{g_j\}_{j=1}^{m}$, we formulate a bipartite graph matching problem where nodes represent predicted and ground truth actions respectively, with edge weights $\text{sim}(p_i,g_j)$ computed as the dot product of their word embeddings encoded by M3-Embedding~\cite{chen-etal-2024-m3}. By computing the dot product between each action embedding from $\text{act}_I^p$ and those from $\text{act}_J^g$, we obtain their similarity matrix $\text{SIM}(\text{act}_I^p,\text{act}_J^g) \in \mathbb{R}^{n\times m}$. To avoid matching dissimilar actions, we apply a similarity threshold $\delta$, 
setting edge weights below it to 0. Our goal is to find the optimal one-to-one 
assignment $\hat{A}$ that maximizes total similarity:

\begin{equation}
\hat{A} = \underset{A\in\Omega}{\arg\max} \sum_{(i,j)\in A} \text{SIM}(\text{act}_I^p,\text{act}_J^g)_{i,j},
\end{equation}

where $\Omega$ represents the set of all valid assignments between predictions and ground truths. We solve this matching problem using the Jonker-Volgenant algorithm~\cite{jonkervolgenant} and subsequently define the precision and recall score for the action sequence as follows:

\begin{equation}
P(\text{act}_I^p,\text{act}_J^g) = \frac{1}{n}\sum_{(i,j)\in \hat{A}} \text{SIM}(\text{act}_I^p,\text{act}_J^g)_{i,j},\quad R(\text{act}_I^p,\text{act}_J^g) = \frac{1}{m}\sum_{(i,j)\in \hat{A}} \text{SIM}(\text{act}_I^p,\text{act}_J^g)_{i,j}.
\end{equation}

The F1 score can be calculated as $F1(\text{act}_I^p,\text{act}_J^g) = \frac{2PR}{P+R}$. The precision, recall, and F1 score for attribute lists are calculated in the same manner.

Since videos may contain multiple objects, we first establish one-to-one correspondences at the entity level between predicted and ground truth objects before computing attribute and action F1 scores. We define the similarity between the $I$-th predicted entity and $J$-th ground truth entity as:

\begin{equation}
\text{sim}(p_I,g_J) = F1(\text{attr}_I^p,\text{attr}_J^g) + F1(\text{act}_I^p,\text{act}_J^g) + \text{sim}(\text{name}_I^p,\text{name}_J^g).
\end{equation}

Using the same matching algorithm, we obtain the optimal entity-level assignment $\hat{A}$. The overall metrics for action sequences are then calculated as:

\begin{equation}
P_{\text{overall\_act}} = \frac{1}{N}\sum_{(I,J)\in \hat{A}} P(\text{act}_I^p,\text{act}_J^g), \quad 
R_{\text{overall\_act}} = \frac{1}{M}\sum_{(I,J)\in \hat{A}} R(\text{act}_I^p,\text{act}_J^g).
\end{equation}

The F1 score can be calculated as $F1_{\text{overall\_act}} = \frac{2P_{\text{overall\_act}}R_{\text{overall\_act}}}{P_{\text{overall\_act}} + R_{\text{overall\_act}}}$. The overall precision, recall, and F1 score for attribute lists are computed in the same manner. The final thinking score(Tscore) for the reasoning process combines these metrics with weighted coefficients:

\begin{equation}
\text{Tscore} = 0.6 \times F1_{\text{overall\_act}} + 0.4 \times F1_{\text{overall\_attr}}.
\end{equation}

\subsubsection{LLM-Assistant Caption Scorer}
As the saying goes, "a picture is worth a thousand words", and a video can be described in numerous valid ways. This makes direct comparison between predicted and ground truth captions challenging for scoring. We therefore design multiple scoring dimensions for caption evaluation, ultimately combining them into an overall score. We employ Qwen2.5-72B\cite{yang2024qwen2.5} as our judge model due to its exceptional language understanding capabilities. Our investigation explores two distinct caption scoring approaches: (1) direct rule-based scoring by the LLM, and (2) event coverage computation through LLM-assisted event extraction, detailed as follows:

\subsection*{Completeness-Naturalness Score (CNscore)}
Since the model first identifies key entities and their attributes/actions during reasoning, the final caption should naturally organize these elements. We evaluate this through two metrics:

\begin{equation}
\text{CNscore} = \frac{\text{Completeness}_{\text{score}} + \text{Naturalness}_{\text{score}}}{20},
\end{equation}

where Completeness$_{\text{score}}$ $\in [0,10]$ measures coverage of reasoned elements, and Naturalness$_{\text{score}}$ $\in [0,10]$ assesses linguistic fluency and human-like description quality. The scoring prompt for Qwen2.5-72B is provided in Appendix \ref{prompt_qwen_score}.

\subsection*{Event Score (Escore)}
Naturalness scoring exhibits significant subjectivity and is susceptible to the inherent biases of the judge model, potentially leading to reward hacking\cite{gao2023scaling} where the model optimizes for generating captions that artificially inflate judge scores while substantially deviating from the desired caption quality objectives. To mitigate this, we avoid direct scoring by the judge model. Considering video descriptions comprise sequences of events (who did what), we evaluate the predicted caption based on event coverage:

\begin{equation}
\text{Escore} = 
\begin{cases}
0 & \text{if event\_coverage} < \delta_{1}, \\
0.5 & \text{if } \delta_{1} \leq \text{event\_coverage} < \delta_{2}, \\
1 & \text{if event\_coverage} \geq \delta_{2}.
\end{cases}
\end{equation}

Here, event coverage represents the proportion of ground truth events entailed by the predicted caption, and we employ Qwen2.5-72B as the judge model to determine these entailment relationships. The prompt for Qwen2.5-72B is the same as Tarsier\cite{wang2024tarsier}.

\subsection{Enhancing Video Description Capabilities of Video MLLMs via GRPO}
\textbf{Reward Function.}
The final reward function for GRPO-based training combines multiple scoring components:$
\text{Reward} = \text{Format}\_\text{score} + \text{Tscore}+ \text{Escore}$.

\textbf{Training Data Construction.}
To effectively reward the model's reasoning process, we construct specialized training data containing explicit structured reasoning annotations. Rather than randomly sampling from existing video captioning datasets, we developed a systematic data selection and annotation pipeline to curate videos exhibiting dynamic motions while ensuring the final training set maintains: (1) diverse action categories with balanced distribution across the dataset, and (2) comprehensive annotations that include both final captions and corresponding reasoning process. The complete data curation pipeline is detailed in the Appendix \ref{data_curation}. Through this process, we established a carefully annotated dataset comprising 1.5k training samples for our experiments. Surprisingly, our model demonstrates substantial performance gains in video captioning despite the limited training set size, validating both the efficacy of our data curation strategy and the robustness of the proposed algorithm. In future work, we plan to scale up training with more data to further boost performance.

\section{Experiments}
\label{sec:Experiments}

\subsection{Experiment Setups}
\textbf{Implementation Details.} We employ Qwen2-VL-7B-Instruct \cite{wang2024qwen2vl} as our baseline model. For both supervised fine-tuning (SFT) and reinforcement learning (RL) training, we utilize the Swift framework\cite{Zhao2024SWIFT}, and we uniformly sample up to 32 frames for each video and resize each frame to a maximum of 460,800 pixels. All experiments are conducted on 8 H800-80GB GPUs. More implementation details are provided in Appendix \ref{more_details}.

\textbf{Evaluation Benchmarks.} We evaluate our model on three video captioning benchmarks: DREAM-1K \cite{wang2024tarsier}, VDC \cite{chai2025auroracap}, and CAREBENCH \cite{xu2024carebench}. DREAM-1K is specifically designed to assess fine-grained action and event description capabilities, featuring dynamic and diverse video content with human-written reference captions. The VDC benchmark comprises over 1,000 videos with exceptionally detailed captions, enabling rigorous evaluation of detailed video description quality. For this benchmark, we employ the official VDCSCORE metric to assess the detailed captioning subtask. CAREBENCH provides comprehensive evaluation of both static objects(spatial elements) and dynamic actions(temporal elements) in captions. To ensure fair comparison, we strictly adhere to the experimental settings specified in each benchmark.

\subsection{Main Results and Analysis}
We conduct comprehensive evaluations of VideoCap-R1 across three established benchmarks, comparing against both general video MLLMs and specialized captioning models (Table \ref{tab:main_results}). VideoCap-R1 demonstrates substantial improvements over the Qwen2-VL-7B baseline even with limited training samples(1.5k), achieving gains of \textbf{+4.4} event F1 on DREAM-1K, \textbf{+4.2} accuracy on VDC, \textbf{+3.1} action F1 and \textbf{+6.9} object F1 on CAREBENCH. Furthermore, our model outperforms all general MLLMs and specialized captioning models by significant margins on both VDC and CAREBENCH. While showing marginally lower event F1 (-0.4\%) than Tarsier-7B on DREAM-1K, VideoCap-R1 exhibits superior performance on the other two benchmarks, indicating stronger generalization capabilities. Notably, VideoCap-R1 achieves a state-of-the-art object F1 score of 34.3\% on CAREBENCH, surpassing even the proprietary GPT-4o-mini (33.8\%), and outperforms Gemini-1.5-Pro by 0.7\% on VDC. These results validate that our structured reasoning approach significantly enhances both precision and recall in describing video entities and actions, ultimately improving overall caption quality.

% \usepackage{booktabs}
% \usepackage{multirow}
\begin{table}[]
\caption{\textbf{Evaluation results on DREAM-1K,VDC and CAREBENCH.} Cells with * are reproduced using the official code. The remaining are reported numbers from literature. We highlight the \textbf{best} results in bold and \underline{second-best} results with underlining. }
\label{tab:main_results}
\centering
\resizebox{.95\textwidth}{!} {
\begin{tabular}{@{}lccccc}
\toprule
\multicolumn{1}{l|}{Model}                                  & \multicolumn{1}{c|}{DREAM-1K}                                                         & \multicolumn{2}{c|}{VDC}                                                                                    & \multicolumn{2}{c}{CAREBENCH}                                                       \\ \cmidrule(l){2-6} 
\multicolumn{1}{l|}{}                                       & \multicolumn{1}{c|}{Event F1/P/R}                                                     & Acc.                           & \multicolumn{1}{c|}{Score}                                                 & Action F1/P/R                            & Object F1/P/R                            \\ \midrule
\textit{Proprietary models}                                 &                                                                                       &                                &                                                                            &                                          &                                          \\
\multicolumn{1}{l|}{{\color[HTML]{9B9B9B} Gemini-1.5-Pro\cite{team2024gemini}}}  & \multicolumn{1}{c|}{{\color[HTML]{9B9B9B} 36.2/37.6/34.8}}                            & {\color[HTML]{9B9B9B} 43.1}    & \multicolumn{1}{c|}{{\color[HTML]{9B9B9B} 2.2}}                            & {\color[HTML]{9B9B9B} -}                 & {\color[HTML]{9B9B9B} -}                 \\
\multicolumn{1}{l|}{{\color[HTML]{9B9B9B} GPT-4o\cite{hurst2024gpt}}}          & \multicolumn{1}{c|}{{\color[HTML]{9B9B9B} 39.2/43.4/35.7}}                            & {\color[HTML]{9B9B9B} -}       & \multicolumn{1}{c|}{{\color[HTML]{9B9B9B} -}}                              & {\color[HTML]{9B9B9B} -}                 & {\color[HTML]{9B9B9B} -}                 \\
\multicolumn{1}{l|}{{\color[HTML]{9B9B9B} GPT-4o mini\cite{hurst2024gpt}}}     & \multicolumn{1}{c|}{{\color[HTML]{9B9B9B} -}}                                         & {\color[HTML]{9B9B9B} -}       & \multicolumn{1}{c|}{{\color[HTML]{9B9B9B} -}}                              & {\color[HTML]{9B9B9B} 36.8/50.2/29.1}    & {\color[HTML]{9B9B9B} 33.8/49.1/25.8}    \\ \midrule
\textit{Open-source models ($>$10B)}                        &                                                                                       &                                &                                                                            &                                          &                                          \\
\multicolumn{1}{l|}{{\color[HTML]{9B9B9B} LLaVA-OV-72B\cite{li2024llava}}}    & \multicolumn{1}{c|}{{\color[HTML]{9B9B9B} 33.2/35.9/30.9}}                            & {\color[HTML]{9B9B9B} -}       & \multicolumn{1}{c|}{{\color[HTML]{9B9B9B} -}}                              & {\color[HTML]{9B9B9B} -}                 & {\color[HTML]{9B9B9B} -}                 \\
\multicolumn{1}{l|}{{\color[HTML]{9B9B9B} LLaVA-Video-72B\cite{zhang2024llavavideo}}} & \multicolumn{1}{c|}{{\color[HTML]{9B9B9B} 34.0/37.3/31.3}}                            & {\color[HTML]{9B9B9B} -}       & \multicolumn{1}{c|}{{\color[HTML]{9B9B9B} -}}                              & {\color[HTML]{9B9B9B} -}                 & {\color[HTML]{9B9B9B} -}                 \\
\multicolumn{1}{l|}{{\color[HTML]{9B9B9B} InternVL2.5-78B\cite{InternVL2.5}}} & \multicolumn{1}{c|}{{\color[HTML]{9B9B9B} 28.6/35.7/23.9}}                            & {\color[HTML]{9B9B9B} -}       & \multicolumn{1}{c|}{{\color[HTML]{9B9B9B} -}}                              & {\color[HTML]{9B9B9B} 28.2/46.4/20.3}    & {\color[HTML]{9B9B9B} 30.5/39.5/24.8}    \\
\multicolumn{1}{l|}{{\color[HTML]{9B9B9B} Qwen2-VL-72B\cite{wang2024qwen2vl}}}    & \multicolumn{1}{c|}{{\color[HTML]{9B9B9B} 33.2/37.3/29.9}}                            & {\color[HTML]{9B9B9B} -}       & \multicolumn{1}{c|}{{\color[HTML]{9B9B9B} -}}                              & {\color[HTML]{9B9B9B} 30.5/47.1/22.6}    & {\color[HTML]{9B9B9B} 24.2/51.9/15.8}    \\ \midrule
\textit{Open-source general MLLMs ($<$10B)}                 &                                                                                       &                                &                                                                            &                                          &                                          \\
\multicolumn{1}{l|}{LLaVA-OV-7B\cite{li2024llava}}                            & \multicolumn{1}{c|}{31.7/34.3/29.4}                                                   & 41.2                           & \multicolumn{1}{c|}{2.1}                                                   & -                                        & -                                        \\
\multicolumn{1}{l|}{LLaVA-Video-7B\cite{zhang2024llavavideo}}                         & \multicolumn{1}{c|}{32.5/37.9/28.4}                                                   & 35.0                           & \multicolumn{1}{c|}{1.8}                                                   & -                                        & -                                        \\
\multicolumn{1}{l|}{InternVL2.5-8B\cite{InternVL2.5}}                         & \multicolumn{1}{c|}{27.6/34.7/22.9}                                                   & \underline{43.0}                           & \multicolumn{1}{c|}{\underline{2.2}}                                                   & 26.0/43.2/18.6                           & 29.1/38.2/23.5                           \\ \midrule
\textit{Open-source specialized captioning MLLMs ($<$10B)}  &                                                                                       &                                &                                                                            &                                          &                                          \\
\multicolumn{1}{l|}{Tarsier-7B\cite{wang2024tarsier}}                             & \multicolumn{1}{c|}{\textbf{34.6}/40.3/30.2}                                                   & 38.3*                              & \multicolumn{1}{c|}{2.1*}                                                     & 27.1/51.1/18.4                           & \underline{31.1}/46.5/23.4                           \\
\multicolumn{1}{l|}{ShareGPT4Video-8B\cite{chen2024sharegpt4video}}                      & \multicolumn{1}{c|}{20.4/27.6/16.1*}                                                  & 35.6                           & \multicolumn{1}{c|}{1.8}                                                   & 16.5/32.6/11.0*                                        & 20.4/42.1/13.4*                                        \\
\multicolumn{1}{l|}{Vriptor\cite{yang2024vript}}                                & \multicolumn{1}{c|}{24.4/23.6/25.1*}                                                  & 38.5                           & \multicolumn{1}{c|}{2.0}                                                   & 23.6/48.7/15.6*                                        & 25.3/38.1/18.9*                                        \\
\multicolumn{1}{l|}{AuroraCap-7B\cite{chai2025auroracap}}                           & \multicolumn{1}{c|}{20.8/24.4/18.1*}                                                  & 41.3                           & \multicolumn{1}{c|}{2.1}                                                   & 21.5/40.3/14.7*                                        & 26.6/34.4/21.6*                                        \\ \midrule
\multicolumn{1}{l|}{Qwen2-VL-7B\cite{wang2024qwen2vl}}                           & \multicolumn{1}{c|}{29.8/33.6/26.8*}                                                   & 39.6*                           & \multicolumn{1}{c|}{2.1*}                                                   & \underline{31.3}/49.5/22.9*                           & 27.4/50.7/18.8*                           \\
\rowcolor[HTML]{EFEFEF} 
\multicolumn{1}{l|}{\cellcolor[HTML]{EFEFEF}\textbf{VideoCap-R1(Ours)}}           & \multicolumn{1}{c|}{\cellcolor[HTML]{EFEFEF}\underline{34.2}$\textcolor{blue}{(+4.4)}$/33.6/34.7} & \textbf{43.8}$\textcolor{blue}{(+4.2)}$ & \multicolumn{1}{c|}{\cellcolor[HTML]{EFEFEF}\textbf{2.4}$\textcolor{blue}{(+0.3)}$} & \textbf{34.4}$\textcolor{blue}{(+3.1)}$/48.2/26.8 & \textbf{34.3}$\textcolor{blue}{(+6.9)}$/50.6/25.9 \\ \bottomrule
\end{tabular}
}
\end{table}

\subsubsection{Superiority of RL to SFT}

Both supervised fine-tuning (SFT) and reinforcement learning (RL) are widely adopted post-training techniques for MLLMs. We investigate their respective impacts on model generalization and reasoning capabilities using identical training data (Table \ref{tab:ablation_sft}). 

The SFT-trained model shows notable gains on DREAM-1K and CAREBENCH, indicating improved action/object description capabilities. However, its generated captions lack detailed attributes and contextual information, resulting in no improvement on the more demanding VDC benchmark. When we introduce structured thinking into the SFT data (Row 3), the model exhibits degraded average performance compared to standard SFT. This suggests SFT's teacher-forcing paradigm merely encourages pattern imitation without establishing genuine reasoning-caption relationships, thereby failing to benefit from the structured thinking process.

In contrast, our GRPO-based two-stage generation strategy demonstrates consistent advantages over SFT across all benchmarks, as the RL framework's inherent self-exploration mechanism coupled with dual think-and-answer rewards enables authentic task decomposition through structured reasoning, ultimately yielding higher-quality descriptions. These experimental results demonstrate that reinforcement learning achieves superior efficacy over supervised fine-tuning for open-ended video captioning tasks, significantly enhancing both model generalization and reasoning capabilities.

% \usepackage{booktabs}
\begin{table}[]
\caption{\textbf{Comparison between SFT and RL.} Our GRPO-based two-stage generation strategy demonstrates consistent advantages over SFT across all benchmarks.}
\label{tab:ablation_sft}
\centering
\resizebox{.95\textwidth}{!} {
\begin{tabular}{l|cccc|l}
\toprule
Model         & DREAM-1K(F1)  & VDC(Acc.)     & CARE-Action(F1) & CARE-Object(F1) & AVG           \\ \midrule
Baseline      & 29.8          & 39.6          & 31.3            & 27.4            & 32.0          \\
+SFT          & 32.8          & 39.8          & 32.6            & 31.5            & 34.2          \\
+SFT with structured thinking & 32.2          & 40.5          & 31.4            & 26.8            & 32.7          \\ \midrule
\rowcolor[HTML]{EFEFEF} 
\textbf{VideoCap-R1(Ours)}          & \textbf{34.2} & \textbf{43.8} & \textbf{34.4}   & \textbf{34.3}   & \textbf{36.7} \\ \bottomrule
\end{tabular}
}
\end{table}

\subsection{Ablation Study}

% \input{tables/ablation_reward}

% \usepackage{booktabs}
\begin{table}[]
\caption{\textbf{Ablation Study on Caption Reward Modeling.} A combination of think score and caption score yields significant boost in performance.}
\label{tab:ablation_2}
\centering
\resizebox{.95\textwidth}{!} {
\begin{tabular}{l|c|cc|cccc|c}
\toprule
Model                                               & Think Score   & \multicolumn{2}{c|}{Caption Score} & DREAM-1K      & VDC           & CAREBENCH     & CAREBENCH     & AVG           \\
                                                    & Tscore        & CNscore          & Escore          & Event F1      & Acc.          & Action F1     & Object F1     &               \\ \midrule
Baseline                                            &               &                  &                 & 29.8          & 39.6          & 31.3          & 27.4          & 32.0          \\ \midrule
w/o Caption Score                                   & $\usym{2714}$ &                  &                 & \textbf{34.3} & 40.3          & 31.4          & 28.3          & 33.6          \\
w/o Think Score                                     &               &                  & $\usym{2714}$   & 30.6          & 43.4          & 33.1          & 33.2          & 35.1          \\ \midrule
\rowcolor[HTML]{EFEFEF} 
\cellcolor[HTML]{EFEFEF}                            & $\usym{2714}$ & $\usym{2714}$    &                 & 32.5          & \textbf{46.8} & \textbf{35.2} & 31.6          & 36.5          \\
\rowcolor[HTML]{EFEFEF} 
\multirow{-2}{*}{\cellcolor[HTML]{EFEFEF}Two Score} & $\usym{2714}$ &                  & $\usym{2714}$   & 34.2          & 43.8          & 34.4          & \textbf{34.3} & \textbf{36.7} \\ \bottomrule
\end{tabular}
}
\end{table}

To validate our caption reward framework, we conduct an ablation study by systematically disabling individual reward components (Table \ref{tab:ablation_2}). Using either think score or caption score alone improves model performance, while their combination yields optimal results, confirming our reward design's effectiveness. The bottom two rows compare our two caption scoring variants: CNscore achieves the highest performance on VDC, whereas Escore delivers the best overall performance across all three benchmarks. Notably, while CNscore leverages LLM-based direct assessment, it suffers from reward hacking - the model tends to generate psychologically nuanced descriptions that appeal to the judge (Qwen2.5-72B) but lack objective video content relevance. This phenomenon explains its suboptimal generalization. We therefore adopt Escore as our default configuration, as its event-based evaluation provides more objective scoring of factual video descriptions, ultimately producing models with superior overall capability.

\subsection{Training Dynamics}

\begin{figure}
    % \vspace{-0.2cm}
    \centering
    \includegraphics[width=0.95\linewidth]{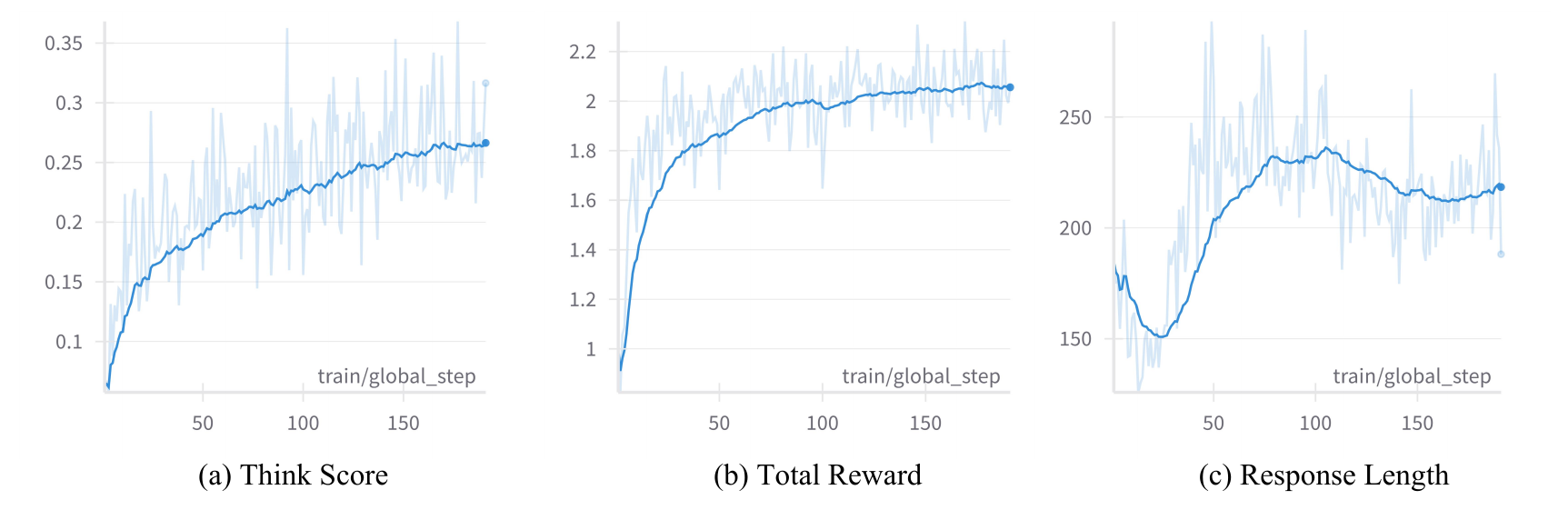}
    \caption{The metric curves of think score, total reward and response length of \textbf{VideoCap-R1}, which show the dynamics of RL training.}
    \label{fig:training_dynamics}
\end{figure}

We primarily monitor the reinforcement learning process using three metrics: think score, total reward, and response length, as illustrated in Figure \ref{fig:training_dynamics}. The think score and total reward exhibit stable increasing trends, demonstrating that the model successfully learns to first perform structured reasoning before generating complete video descriptions, which validates the effectiveness of our carefully designed reward signals. 

The response length of VideoCap-R1 initially decreases, then increases before relatively stabilizing - the early-phase pattern aligns with observations from prior work \cite{zeng2025simplerl}. This trajectory indicates that the RL training progressively replaces the model's original reasoning patterns with the new reasoning style. Notably, the response length does not grow indefinitely but shows slight fluctuations, which we attribute to varying information density across videos. This length variation suggests the model learns to adapt its description length according to the actual video content.
\definecolor{Green}{rgb}{0,0.7,0.3}
\definecolor{HighlightGreen}{rgb}{0.88,0.94,0.85}
\subsection{Qualitative Results}
\begin{figure}
    \centering
    \includegraphics[width=0.95\linewidth]{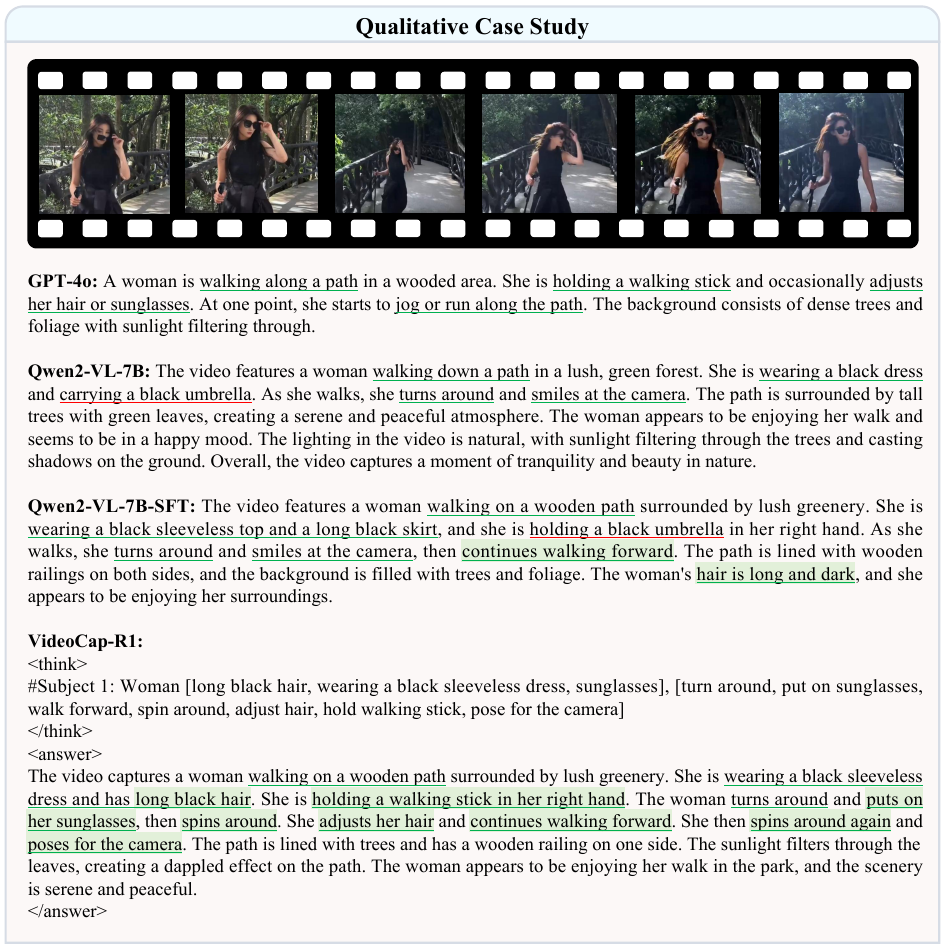}
    \caption{Qualitative comparison between VideoCap-R1 and baseline models. We annotate examples with: \setulcolor{red}{\ul{red underlines}} for hallucinated content, \setulcolor{Green}{\ul{green underlines}} for accurate descriptions, and \colorbox{HighlightGreen}{highlighting} for correct details uniquely captured by VideoCap-R1 or Qwen2-VL-7B-SFT (vs. Qwen2-VL-7B). Our model demonstrates superior fine-grained content description capabilities. }
    \label{fig:qualitative_comparison1}
\end{figure}
Figure \ref{fig:qualitative_comparison1} presents qualitative comparisons between VideoCap-R1 and baseline models. Both the baseline Qwen2-VL and its SFT-finetuned version erroneously hallucinate the "stick" held by the woman as an "umbrella". In contrast, our VideoCap-R1 correctly identifies the stick while additionally providing: (1) more detailed attributes, (2) finer-grained action descriptions, and (3) even capturing subtle motions (e.g., "spins around") missed by GPT-4o. Notably, VideoCap-R1's structured reasoning content demonstrates strong alignment with final descriptions, evidencing that it genuinely establishes and benefits from the reasoning-description relationship, ultimately enhancing overall caption quality through this cognitive process. See Appendix \ref{qualitative_results} for more comparisons.

\section{Conclusion and Future Work}
\label{sec:Conclusions}
In this work, we have investigated GRPO-based reinforcement learning for post-training video MLLMs to enhance their capability in describing actions and events. Our VideoCap-R1 incorporates three key designs: (1) a two-stage generation strategy with structured thinking, (2) Tscore for rewarding the thinking process, and (3) caption score for evaluating the final descriptions. The proposed model demonstrates significant improvements over baseline approaches, achieving superior performance even with limited training samples (1.5k) and outperforming SFT-trained counterparts across all benchmarks. Future work will focus on scaling up the training data to further enhance video description capabilities through reinforcement learning. We hope VideoCap-R1 can serve as a strong foundation for future research on developing more advanced video captioning systems through reinforcement learning techniques.

%\clearpage
{
\small
\bibliographystyle{plainnat}
\bibliography{reference}
}

\clearpage
\appendix
\section*{Appendix}
This supplementary material includes the following sections:
\begin{itemize}
    \item In Section \ref{more_details}, we provide more implementation details.
    \item In Section \ref{data_curation}, we provide the details of training data curation.
    \item In section \ref{qualitative_results}, we provide more qualitative comparisons between VideoCap-R1 and baseline models.
    \item In section \ref{prompt_qwen_score}, we give the scoring prompt for Qwen2.5-72B.
    \item In section \ref{training_prompt}, we give the training prompt for Qwen2-VL-7B.
\end{itemize}
\section{More Implementation Details}
\label{more_details}
Table \ref{tab:hyper_para} shows the training hyper-parameters in SFT and GRPO. For GRPO optimization, we perform 7 rollouts per prompt($G=7$) and set the sampling temperature to 1.0. We adopt $\beta = 0.001$ for KL penalization and set the thresholds of event coverage $\delta_1=0.28, \delta_2=0.35$. All experiments are conducted on 8 H800-80GB GPUs. For GRPO, we allocate 7 GPUs for training and reserve 1 GPU exclusively for rollouts, while SFT utilizes all 8 GPUs for training.

% \usepackage{booktabs}
% \usepackage{multirow}
\begin{table}[h]
\caption{Training hyper-parameters of VideoCap-R1.}
\label{tab:hyper_para}
\centering
\resizebox{.4\textwidth}{!} {
\begin{tabular}{@{}l|cc@{}}
\toprule
Configuration          & SFT            & GRPO           \\ \midrule
Baseline               & \multicolumn{2}{c}{Qwen2-VL-7B} \\
Optimizer name         & \multicolumn{2}{c}{AdamW}       \\
Optimizer $\beta_1$    & \multicolumn{2}{c}{0.9}         \\
Optimizer $\beta_2$    & \multicolumn{2}{c}{0.999}       \\
Optimizer eps          & 1e-6           & 1e-8           \\
Learning rate          & \multicolumn{2}{c}{1e-6}        \\
Learning rate schedule & \multicolumn{2}{c}{cosine}      \\
Training epoch         & \multicolumn{2}{c}{1}           \\
Warm-up ratio          & 0.05           & 0.01           \\
Weight decay           & 0.01           & 0.1            \\
Global batch size      & 64             & 56              \\ \bottomrule
\end{tabular}
}
\end{table}

\section{Training Data Curation}
\label{data_curation}

\textbf{Dynamic Video Selection.} 
We construct our training set by sampling from the Tarsier2-Recap-585K dataset \cite{yuan2025tarsier2}, as it provides exceptionally accurate and detailed video descriptions with comprehensive action annotations. To ensure the selected videos exhibit sufficient dynamic content for improving action/event description capabilities, we implement an optical-flow-based filtering pipeline that: (1) computes frame-to-frame optical flow intensity as a dynamicity metric, and (2) retains only videos with both high dynamicity scores and appropriate durations (10-30 seconds).

\begin{figure}[ht]
  \centering
  \includegraphics[width=0.8\linewidth]{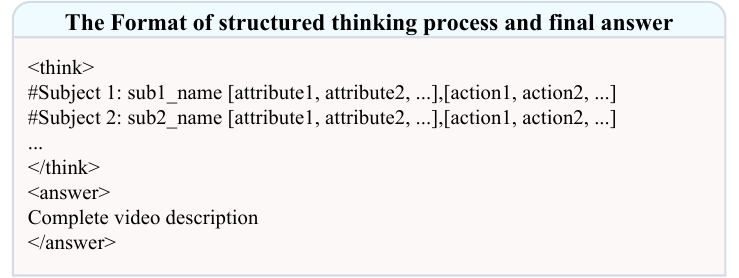}
  \caption{The Format of structured thinking process and final answer.}
  \label{fig:format_think}
\end{figure}

\begin{figure}[ht]
  \centering
  \includegraphics[width=0.85\linewidth]{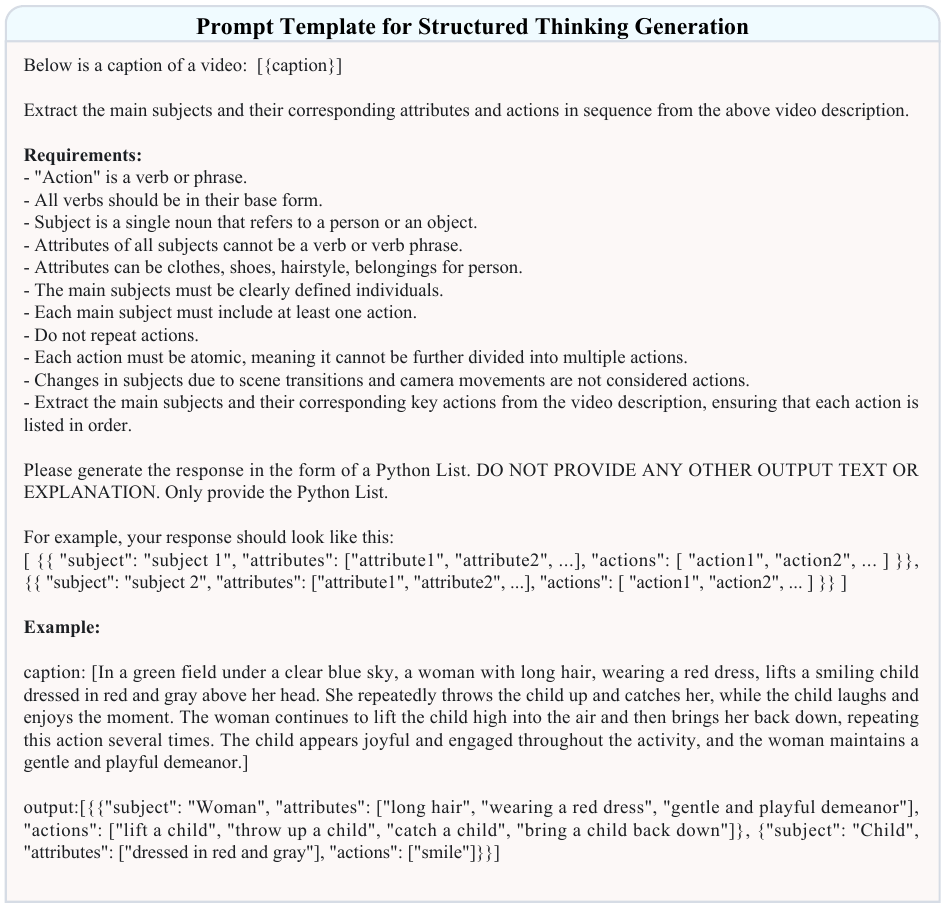}
  \caption{Prompt Template for Structured Thinking Generation.}
  \label{fig:prompt_cot_generation}
\end{figure}

\textbf{Structured Thinking Annotation.} 
To effectively reward the model’s reasoning process(as shown in Figure \ref{fig:format_think}), we construct specialized training data containing explicit structured thinking annotations. We design a carefully engineered prompt template (Figure \ref{fig:prompt_cot_generation}) to guide Qwen2.5-72B \cite{yang2024qwen2.5} in producing structured reasoning content. Each annotation must satisfy: (i) maximal coverage of main subjects while maintaining attribute/action consistency, and (ii) strict temporal alignment between described actions and actual video progression.

\textbf{Action-Based Video Filtering.} 
To ensure balanced action distribution in our training set, we implement an incremental filtering mechanism that maintains a running inventory of covered actions. Videos are selectively added only when they introduce new action types not already represented in our dataset. Through this process, we construct a curated dataset of 1.5K video samples characterized by: (1) high dynamicity, (2) diverse action types, and (3) paired structured reasoning annotations with detailed descriptions.

\section{Qualitative Results}
\label{qualitative_results}
\begin{figure}
    \centering
    \includegraphics[width=0.95\linewidth]{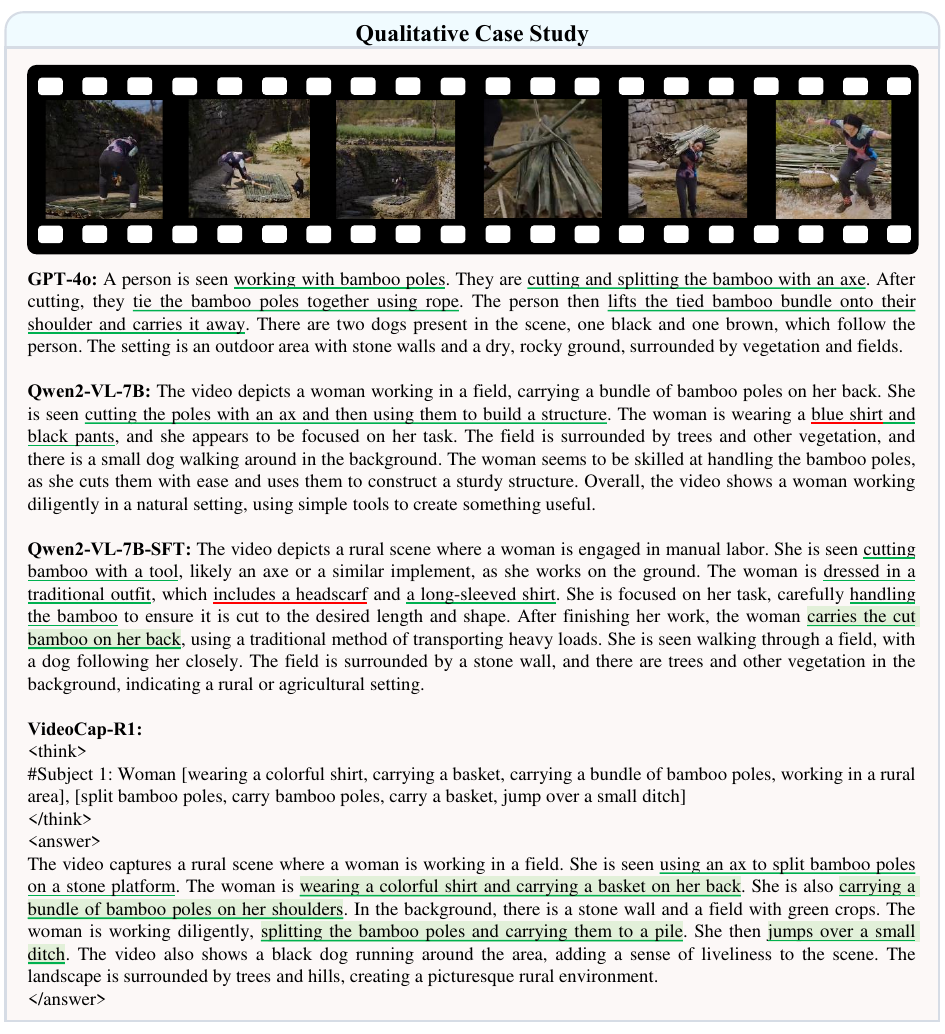}
    \caption{The first qualitative comparison between VideoCap-R1 and baseline models. The video is from DREAM-1K. We annotate examples with: \setulcolor{red}{\ul{red underlines}} for hallucinated content, \setulcolor{Green}{\ul{green underlines}} for accurate descriptions, and \colorbox{HighlightGreen}{highlighting} for correct details uniquely captured by VideoCap-R1 or Qwen2-VL-7B-SFT (vs. Qwen2-VL-7B). Our model demonstrates superior fine-grained content description capabilities. }
    \label{fig:qualitative_comparison2}
\end{figure}

\begin{figure}
    \centering
    \includegraphics[width=0.95\linewidth]{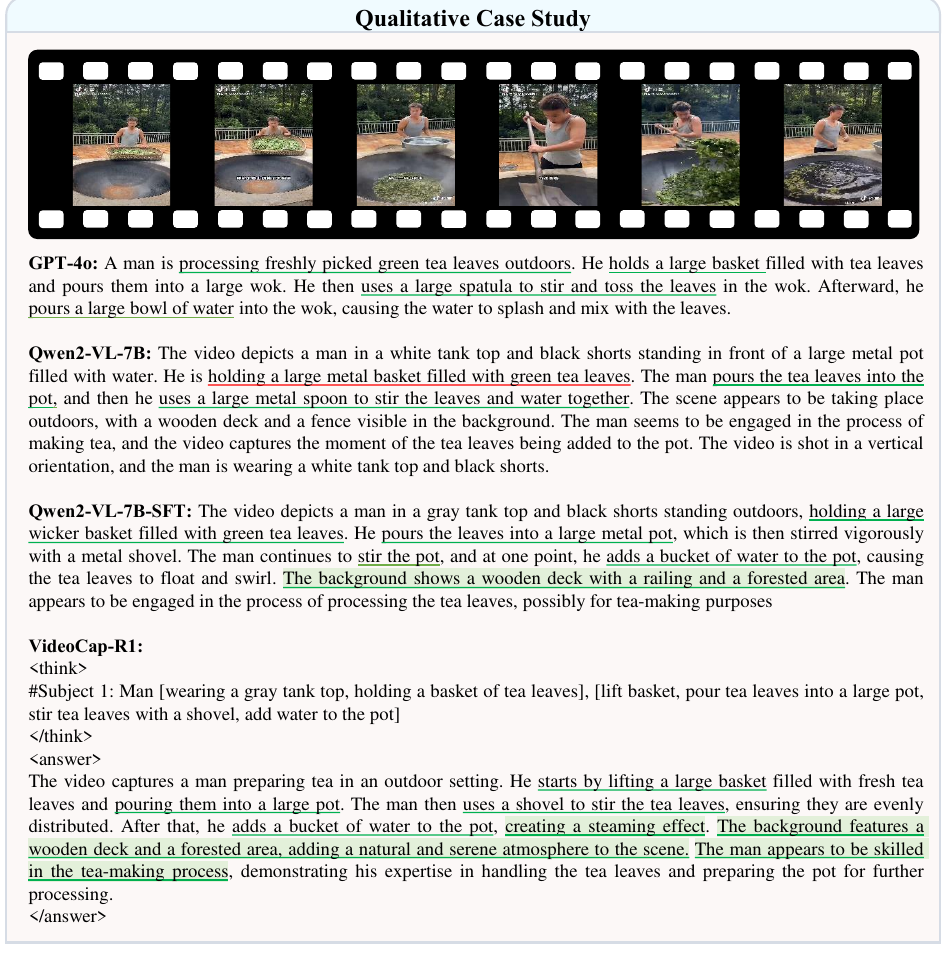}
    \caption{The second qualitative comparison between VideoCap-R1 and baseline models. The video is from DREAM-1K. We annotate examples with: \setulcolor{red}{\ul{red underlines}} for hallucinated content, \setulcolor{Green}{\ul{green underlines}} for accurate descriptions, and \colorbox{HighlightGreen}{highlighting} for correct details uniquely captured by VideoCap-R1 or Qwen2-VL-7B-SFT (vs. Qwen2-VL-7B). Our model demonstrates superior fine-grained content description capabilities. }
    \label{fig:qualitative_comparison3}
\end{figure}

\begin{figure}
    \centering
    \includegraphics[width=0.95\linewidth]{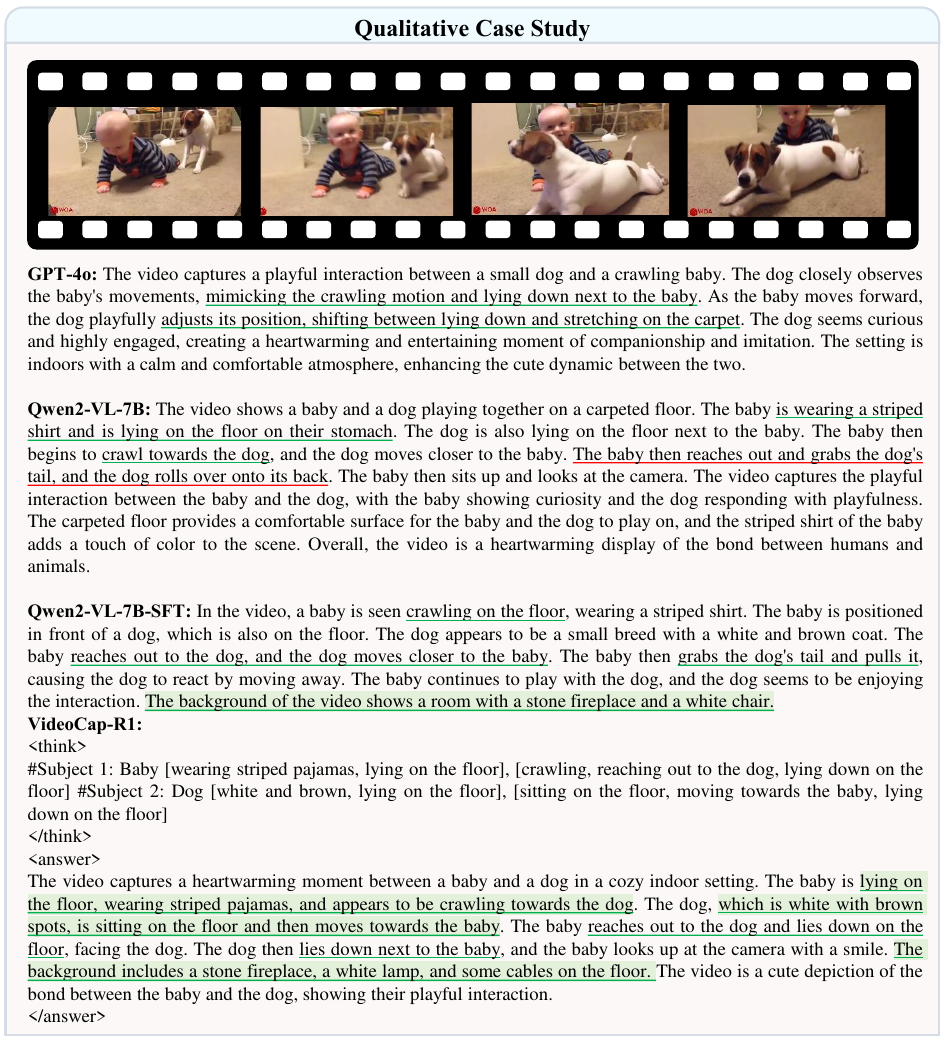}
    \caption{The third qualitative comparison between VideoCap-R1 and baseline models. The video is from CAREBENCH. We annotate examples with: \setulcolor{red}{\ul{red underlines}} for hallucinated content, \setulcolor{Green}{\ul{green underlines}} for accurate descriptions, and \colorbox{HighlightGreen}{highlighting} for correct details uniquely captured by VideoCap-R1 or Qwen2-VL-7B-SFT (vs. Qwen2-VL-7B). Our model demonstrates superior fine-grained content description capabilities. }
    \label{fig:qualitative_comparison4}
\end{figure}

\begin{figure}
    \centering
    \includegraphics[width=0.95\linewidth]{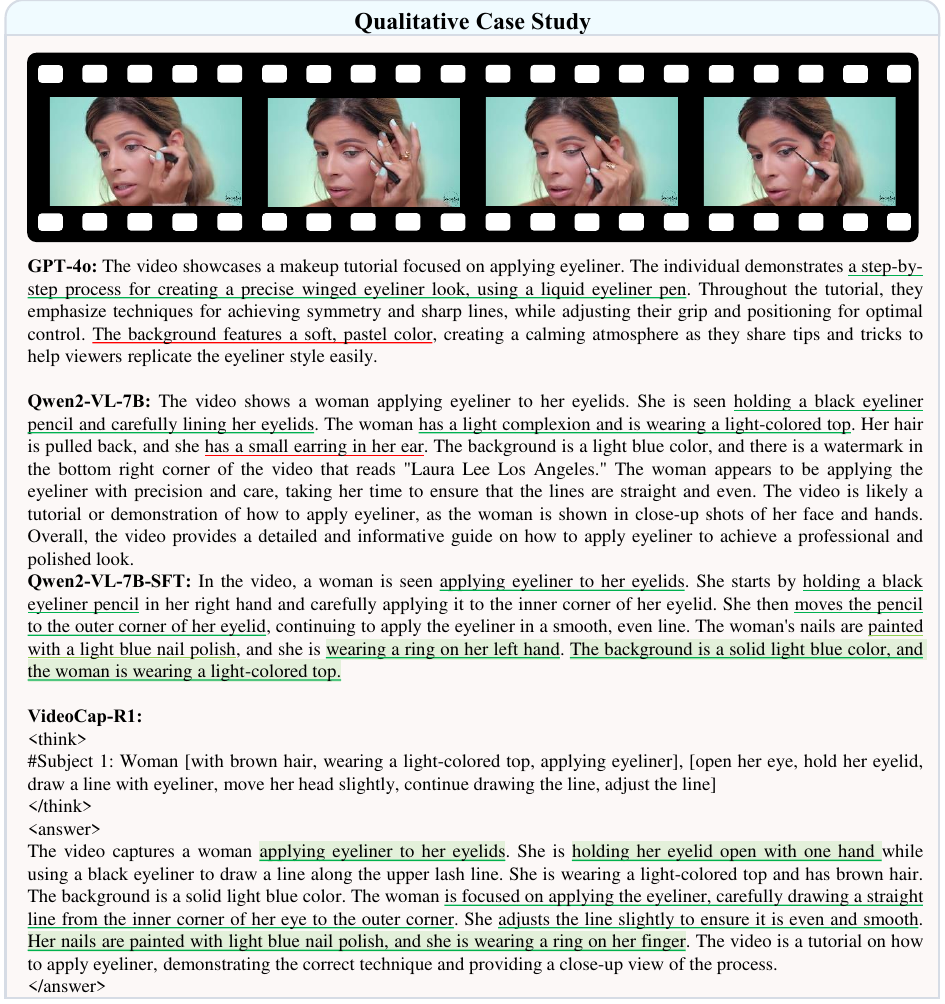}
    \caption{The fourth qualitative comparison between VideoCap-R1 and baseline models. The video is from CAREBENCH. We annotate examples with: \setulcolor{red}{\ul{red underlines}} for hallucinated content, \setulcolor{Green}{\ul{green underlines}} for accurate descriptions, and \colorbox{HighlightGreen}{highlighting} for correct details uniquely captured by VideoCap-R1 or Qwen2-VL-7B-SFT (vs. Qwen2-VL-7B). Our model demonstrates superior fine-grained content description capabilities. }
    \label{fig:qualitative_comparison5}
\end{figure}

\begin{figure}
    \centering
    \includegraphics[width=0.95\linewidth]{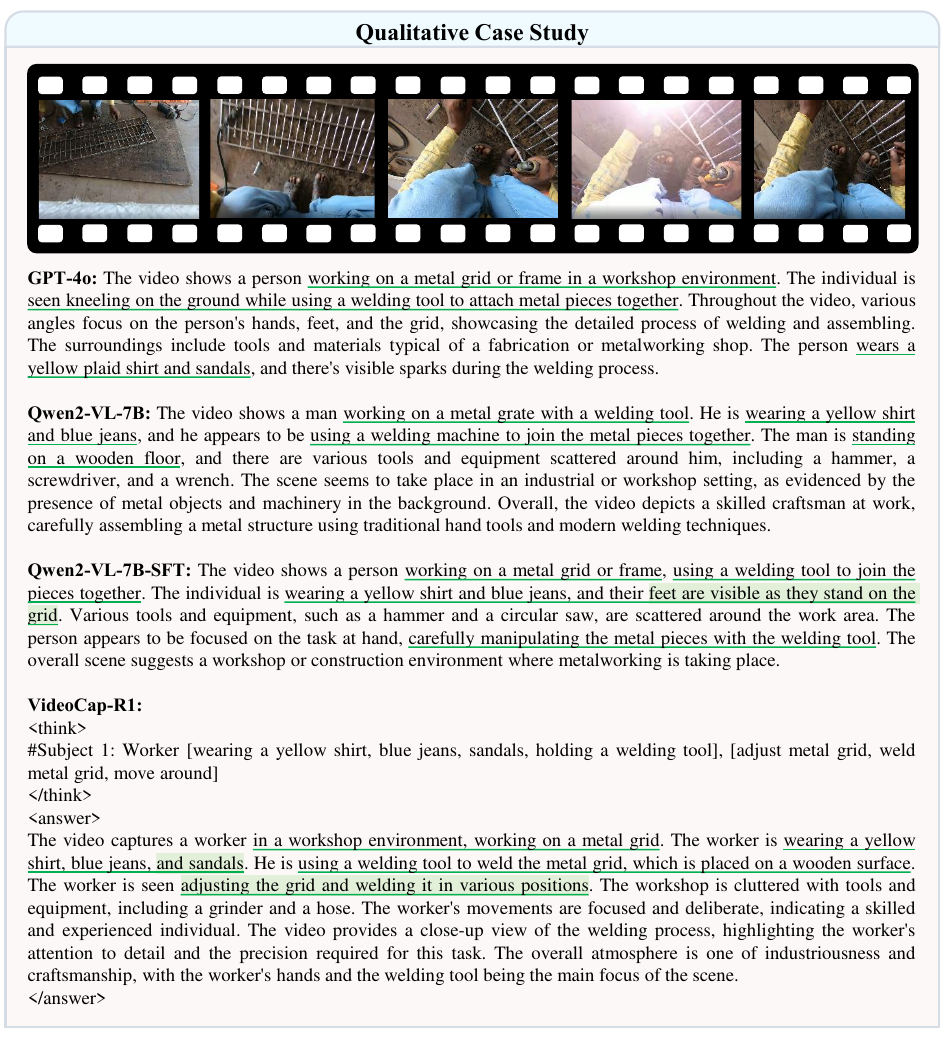}
    \caption{The fifth qualitative comparison between VideoCap-R1 and baseline models. The video is from VDC. We annotate examples with: \setulcolor{red}{\ul{red underlines}} for hallucinated content, \setulcolor{Green}{\ul{green underlines}} for accurate descriptions, and \colorbox{HighlightGreen}{highlighting} for correct details uniquely captured by VideoCap-R1 or Qwen2-VL-7B-SFT (vs. Qwen2-VL-7B). Our model demonstrates superior fine-grained content description capabilities. }
    \label{fig:qualitative_comparison6}
\end{figure}

Figures \ref{fig:qualitative_comparison2}--\ref{fig:qualitative_comparison6} present additional qualitative comparisons between VideoCap-R1 and baseline models on representative samples from the DREAM-1K, VDC, and CAREBENCH benchmarks. These examples consistently demonstrate our model's advantages in: (i) fine-grained action characterization, (ii) precise object attribute description, and (iii) comprehensive event coverage across diverse video content.


\newpage

\section{The scoring prompt for Qwen2.5-72B}
\label{prompt_qwen_score}
\begin{figure}[ht]
  \centering
  \includegraphics[width=0.95\linewidth]{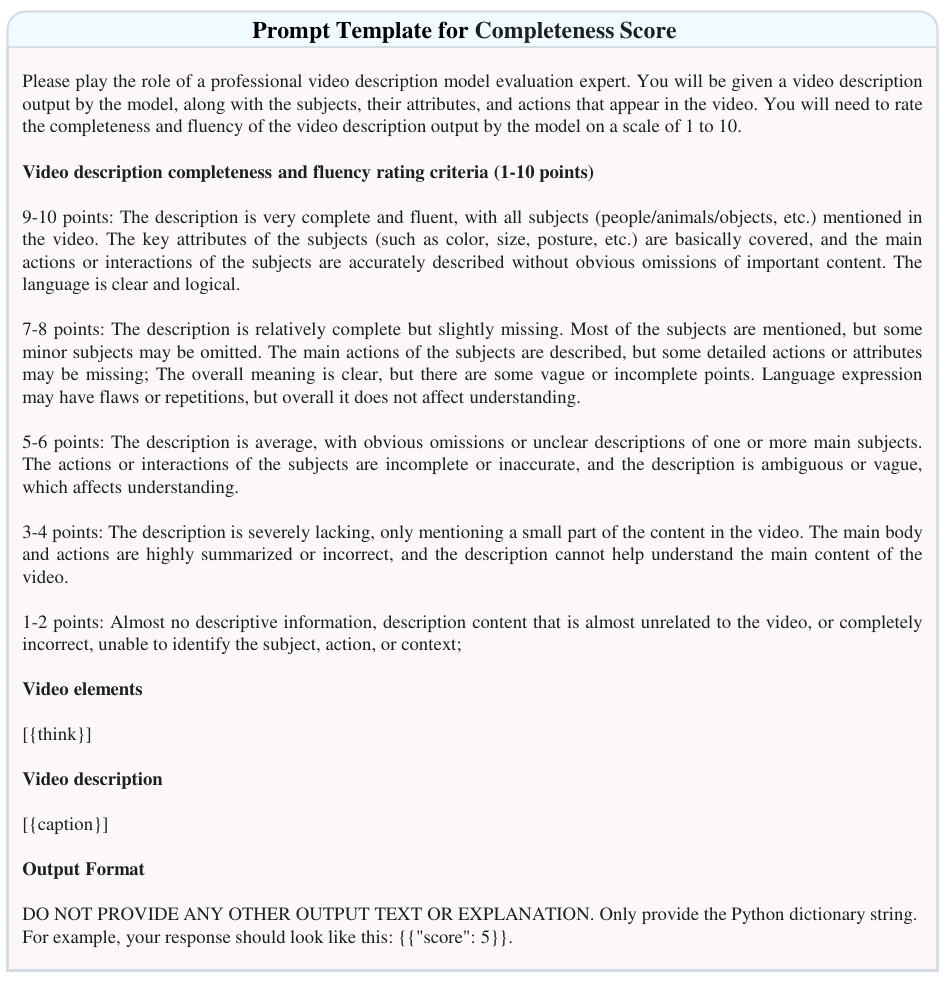}
  \caption{Prompt Template for Completeness Score.}
  \label{fig:cscore}
\end{figure}

\begin{figure}[ht]
  \centering
  \includegraphics[width=0.85\linewidth]{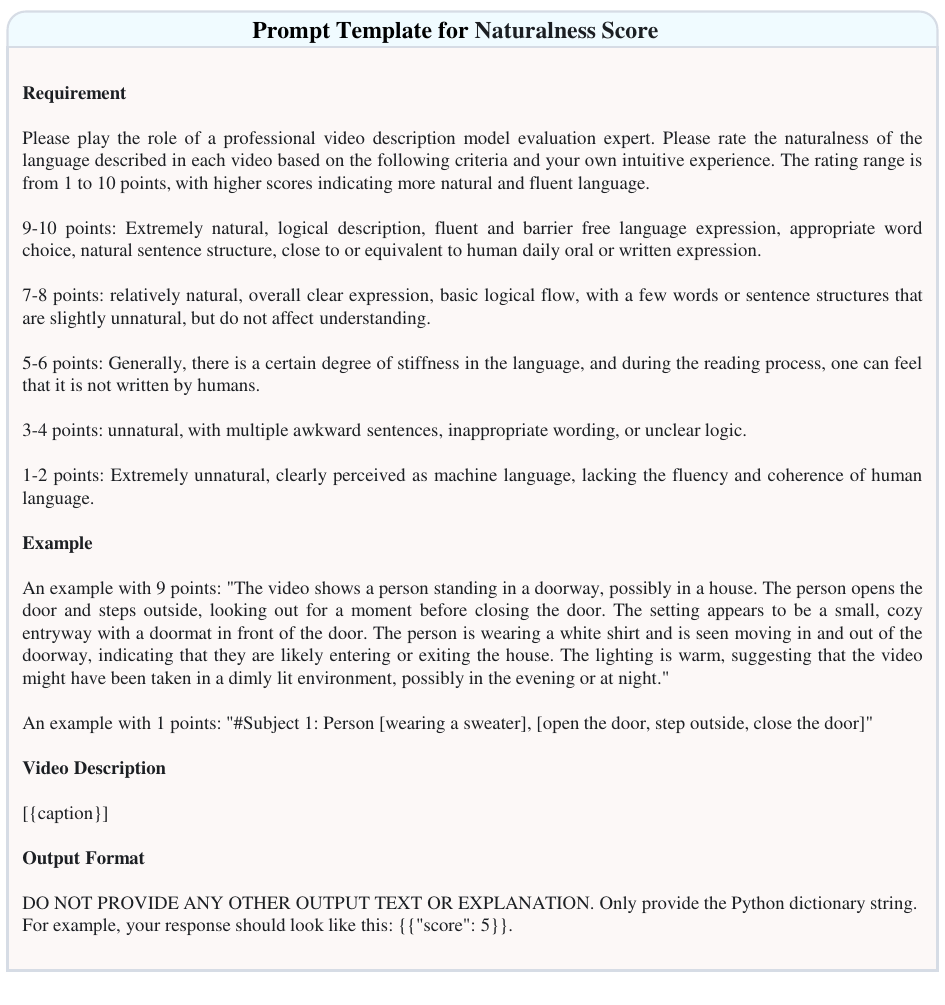}
  \caption{Prompt Template for Naturalness Score.}
  \label{fig:nscore}
\end{figure}

Figures \ref{fig:cscore} and \ref{fig:nscore} present the prompt templates for Qwen2.5-72B to assess caption completeness and naturalness scores, respectively. 

\section{The Training Prompt for Qwen2-VL-7B used in GRPO}
\label{training_prompt}
\begin{figure}[ht]
  \centering
  \includegraphics[width=0.85\linewidth]{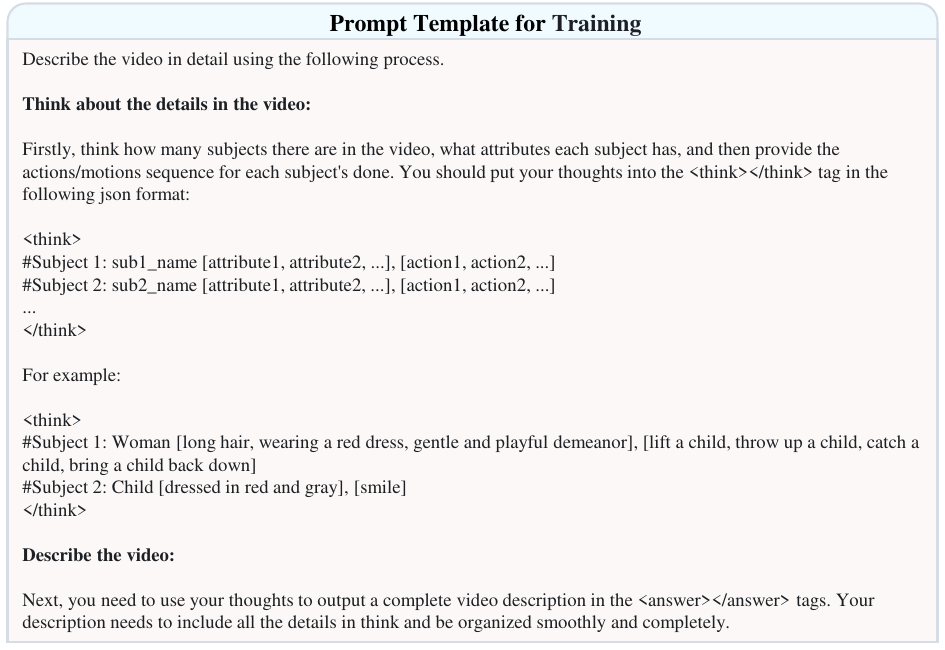}
  \caption{Prompt Template for Training.}
  \label{fig:training_template}
\end{figure}

Figure \ref{fig:training_template} presents the prompt template employed for both GRPO-based reinforcement learning training and supervised fine-tuning (SFT) on our structured-thinking-augmented instruction dataset.

\end{document}